\documentclass[conference]{IEEEtran}
\IEEEoverridecommandlockouts
\usepackage{cite}
\usepackage{amsmath,amssymb,amsfonts}
\usepackage{algorithmic}
\usepackage{graphicx}
\usepackage{textcomp}
\usepackage{xcolor}
\usepackage{enumerate}
\usepackage{array,booktabs,arydshln,xcolor}
\usepackage[hidelinks]{hyperref}
\usepackage{caption}
\usepackage{enumitem}
\usepackage{tablefootnote}
\usepackage{comment}

\newcommand{\cmmnt}[1]{}

\def\BibTeX{{\rm B\kern-.05em{\sc i\kern-.025em b}\kern-.08em
    T\kern-.1667em\lower.7ex\hbox{E}\kern-.125emX}}

\begin{document}

\title{Detecting the Unexpected: AI-Driven Anomaly Detection in Smart Bridge Monitoring \\
}

\author{
    \IEEEauthorblockN{Rahul Jaiswal$^{*}$, Joakim Hellum~and Halvor Heiberg}
    \IEEEauthorblockA{
    Smart Sensor Systems AS \\
    Oslo, Norway \\
    \{rahul.jaiswal, joakim.hellum, halvor.heiberg\}@smartsensorsystems.no}
    \thanks{$^{*}$Corresponding author}
}

\IEEEoverridecommandlockouts \IEEEpubid{\makebox[\columnwidth]{979-8-3315-3672-5/26/\$31.00 \copyright 2026 IEEE \hfill} \hspace{\columnsep}\makebox[\columnwidth]{ }}

\maketitle

\begin{abstract} 
Bridges are critical components of national infrastructure and smart cities. Therefore, smart bridge monitoring is essential for ensuring public safety and preventing catastrophic failures or accidents. Traditional bridge monitoring methods rely heavily on human visual inspections, which are time-consuming and prone to subjectivity and error. This paper proposes an artificial intelligence (AI)-driven anomaly detection approach for smart bridge monitoring. Specifically, a simple machine learning (ML) model is developed using real-time sensor data collected by the iBridge sensor devices installed on a bridge in Norway. The proposed model is evaluated against different ML models. Experimental results demonstrate that the density-based spatial clustering of applications with
noise (DBSCAN)-based model outperforms in accurately detecting the anomalous events (bridge accident). These findings indicate that the proposed model is well-suited for smart bridge monitoring and can enhance public safety by enabling the timely detection of unforeseen incidents.
\end{abstract}

\section{Introduction} 
Bridges are the critical component of smart cities, forming the backbone of modern transportation systems. With hundreds of thousands of bridges deployed worldwide, for instance, over 18,000 in Norway~\cite{nor_ency}, ensuring their safety and reliability is of paramount importance. Traditionally, the assessment of bridge conditions has relied heavily on human visual inspections. However, such inspections are often time-consuming and prone to error, as accurately evaluating the extent of structural deterioration, particularly in concrete/wooden bridges, is challenging due to the limitations of human visual observations. To overcome these challenges, intelligent sensor-based, data-driven machine-learning bridge monitoring techniques have emerged. It enables timely assessment of structural integrity, enhancing pedestrian and public safety, and mitigating the risk of catastrophic failures or accidents. Effective bridge monitoring requires designing a smart model that can capture variations between normal and anomalous sensor data.

\begin{figure} [t!]
  \centering
  \includegraphics[width=\columnwidth, height=4.5cm, keepaspectratio]{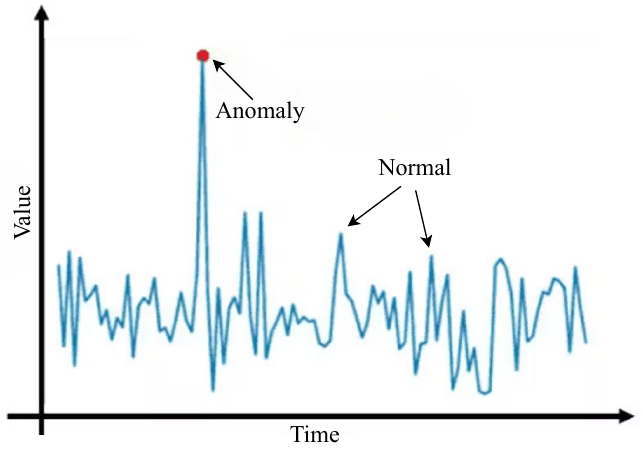}
  \caption{A toy example of anomaly detection in sensor data.}
  \label{anomaly_exm}  \vspace{-4mm}
\end{figure}

Anomaly detection is an important application of data mining~\cite{han2022data} and is commonly referred to as outlier detection. In the context of bridge sensor data, anomalies or outliers correspond to the measurements that deviate from the normal signal or data patterns. These deviations may arise from structural damage, abnormal loading conditions, environmental effects, or sensor malfunctions. The objective of anomaly detection is therefore to identify such inconsistent or irregular sensor observations that differ significantly from typical bridge behaviour. A toy example illustrating anomaly detection in sensor data is presented in Fig.~\ref{anomaly_exm}. As shown, the anomalous data point exhibits a significantly higher peak compared to the normal sensor measurements.

Several approaches have been proposed in the literature for detecting anomalies in sensor data. Statistical-based methods~\cite{samariya2023comprehensive}, such as multivariate models~\cite{edirisinghe2015markov} and mean-variance-based techniques~\cite{elton2009modern}, typically learn discriminative features from historical monitoring data representing normal structural behaviour. A data point is then classified as anomalous when it deviates significantly from these learned features. However, the effectiveness of statistical methods relies on the prior empirical knowledge, including assumptions about data distributions and model parameters used for anomaly identification. In practice, the true distribution of real-world sensor data is often unknown, which can limit the accuracy of anomaly detection using statistical methods.

Recently, artificial intelligence (AI)-driven, particularly machine learning (ML)-based, approaches have gained significant attention for anomaly detection~\cite{jaiswal2025leveraging}, as they automatically extract discriminative features from raw bridge sensor data to train models. Neural network-based methods are also effective due to their strong representation and universal approximation capabilities~\cite{jaiswal2018sound,jaiswal2023caqoe}. For instance, a K-nearest neighbour-based anomaly detection framework is proposed in~\cite{entezami2020big} using raw vibration measurements collected from a bridge. A convolutional neural network (CNN) is used in~\cite{abdeljaber2017real} to detect structural damages in bridge systems. An autoencoder is employed in~\cite{tian2025anomaly} for anomaly detection in bridge structural health monitoring. 

The review of existing literature highlights that anomaly detection in bridge systems is critical for maintaining the structural performance of the bridge, ensuring pedestrian and public safety, and mitigating the risk of catastrophic failures or accidents. Motivated by these findings, this work focuses on machine learning-based anomaly detection for smart bridge monitoring. Specifically, we develop an efficient ML-based anomaly detection framework to identify a bridge accident captured by iBridge sensor devices (see Section~\ref{data_coll}) installed on a bridge in Norway. The proposed model can enable bridge monitoring authorities to promptly detect anomalous events and take timely corrective actions, thereby reducing potential risks and preventing further adverse consequences.

The remainder of this paper is organized as follows. Section~\ref{ml_bgd} describes the machine learning methods employed for anomaly detection. Section~\ref{prop_model} introduces the proposed anomaly detection model. Section~\ref{dset} describes the experimental dataset. Section~\ref{res_dis} discusses the results, and Section~\ref{cons} concludes the paper and outlines directions for future work.

\section{Background} 
\label{ml_bgd}
This section describes various machine learning techniques used in our experimental study for detecting anomalies in the bridge monitoring system. 

\subsection{Isolation Forest}
\label{if_algo}
\begin{figure} [t!]
  \centering
  \includegraphics[width=\columnwidth, height=5.25cm, keepaspectratio]{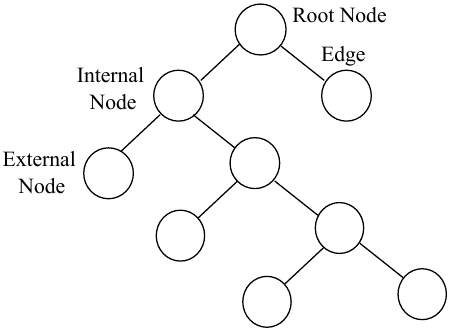}
  \caption{A simple illustration of an iTree structure.}
  \label{IF_tree} 
\end{figure}

The Isolation Forest (IF) algorithm detects anomalies using an ensemble of binary trees~\cite{chabchoub2022depth}. In bridge monitoring applications, sensor data are isolated by randomly selecting a feature and an associated split value at each node. The algorithm constructs a forest, referred to as an iForest, comprising multiple isolation trees (iTrees)~\cite{xu2023deep}. During training, the sensor data are recursively partitioned until the iTrees effectively separate normal data points from anomalous data points. Typically, anomalous data points are isolated near the root nodes of the iTrees, whereas normal operating conditions (points) require deeper traversals and are located farther from the root nodes. A simplified structure of an iTree~\cite{jaiswal2025leveraging} is illustrated in Fig.~\ref{IF_tree}. Here, external nodes have no children, whereas internal nodes contain two child nodes. In this representation, anomalies are typically associated with external nodes. 

\subsection{Autoencoder}
\label{ae_algo}
An autoencoder (AE)~\cite{abhaya2023efficient} is a neural network designed to learn a compact (latent) representation of input data while minimizing the error in reconstructing the original signal. In bridge monitoring applications, autoencoders are typically trained using sensor measurement data that capture normal signal behaviour. Consequently, when anomalous sensor measurements, such as those caused by structural damage, a vehicular accident, an unusual load on the bridge, or sensor malfunctions, are encountered, the reconstruction error increases significantly. This reconstruction error can therefore be used as an effective indicator for anomaly detection.

\begin{figure} [t!]
  \centering
  \includegraphics[width=\columnwidth, height=5.25cm, keepaspectratio]{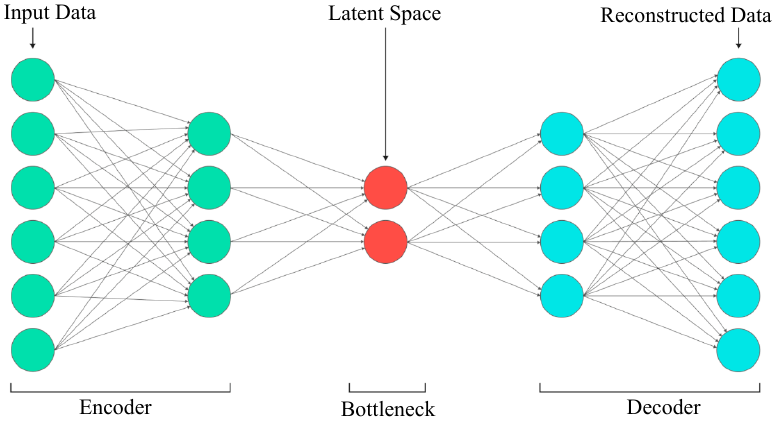}
  \caption{A simple illustration of an autoencoder.}
  \label{ae_design} 
\end{figure}

An autoencoder consists of an encoder and a decoder. The encoder compresses input data into a low-dimensional latent representation (also called a bottleneck), and the decoder reconstructs the original input from that latent representation, as shown in Fig.~\ref{ae_design}. The autoencoder model is trained by minimizing the reconstruction error between the input and its reconstruction. Next, we briefly present the mathematical formulation of an autoencoder.

Let a sensor device with $d$ channels generate measurements over time. At time $t$, the sensor measurements are given as:
\begin{equation}
\mathbf{x}_t = x_t^{(1)}, x_t^{(2)}, \dots, x_t^{(d)}
\end{equation}
where, $x_t^{(i)}$ denotes the measurements of the i-th sensor device.

The encoder compresses the input into a low-dimensional latent representation as:
\begin{equation}
\mathbf{z} = f_{\theta}(\mathbf{x})
           = \sigma\left(\mathbf{W}_{e}\mathbf{x} + \mathbf{b}_{e}\right)
\end{equation}

where, $\mathbf{W}_{e} \in \mathbb{R}^{k \times D}$ is encoder weight matrix,$\quad
\mathbf{b}_{e} \in \mathbb{R}^{k}$ is the encoder bias vector, $\quad k < D$ is the dimension of the latent space which controls the compression strength, $D$ is the dimension of the input vector, $\mathbf{z} \in \mathbb{R}^{k}$ is the latent vector, and $\sigma(\cdot)$ is a nonlinear activation function (e.g., ReLU, tanh). Note that $\mathbb{R}$ denotes the set of real numbers.

The decoder reconstructs the original input from the latent representation as:
\begin{equation}
\hat{\mathbf{x}} = g_{\phi}(\mathbf{z})
                 = \sigma\left(\mathbf{W}_{d}\mathbf{z} + \mathbf{b}_{d}\right)
\end{equation}

where, $\mathbf{W}_{d} \in \mathbb{R}^{D \times k}$ and $\quad \mathbf{b}_{d} \in \mathbb{R}^{D}$.

Next, the reconstruction error measures how well the model reproduces sensor signal patterns, and it is given by the mean squared error (MSE)~\cite{wang2009mean} as:
\begin{equation}
\mathcal{L}(\mathbf{x}, \hat{\mathbf{x}})
=
\frac{1}{D}
\sum_{i=1}^{D}
\left(x_i - \hat{x}_i\right)^2
\end{equation}

To determine whether a given sample is normal or anomalous, an anomaly score based on reconstruction error (MSE) is computed and compared against a predefined threshold. If the anomaly score is below the threshold, the sample is classified as normal; otherwise, it is classified as anomalous. Note that normal samples exhibit low reconstruction error, whereas anomalous samples result in high reconstruction error.

\subsection{Density-based Spatial Clustering of Application with Noise}
\label{dbscan_algo}
The Density-based Spatial Clustering of Applications with Noise (DBSCAN)~\cite{chien2024density} is a density-based clustering technique that identifies data patterns by analyzing the local concentration of samples. In bridge monitoring applications, DBSCAN assumes that sensor measurements corresponding to normal data points form dense regions in the feature space, whereas abnormal events such as accidents or unusual loads appear as isolated points with low local density.

The method relies on two key parameters. The neighbourhood radius \textit{eps}, which defines the maximum distance between two sensor measurements for them to be considered neighbours and form a cluster. The minimum number of points \textit{minpts}, which specifies the minimum number of neighbouring measurements required within the \textit{eps} radius for a data point to be considered as part of the cluster~\cite{chien2024density}.

\begin{figure} [b!]
  \centering
  \includegraphics[width=\columnwidth, height=5.0cm, keepaspectratio]{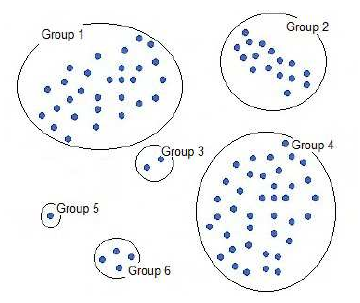}
  \caption{A simple illustration of the DBSCAN method.}
  \label{dbscan_design} \vspace{-3mm}
\end{figure}

During training, DBSCAN categorizes samples into three groups. The samples having a sufficient number of neighbours within the \textit{eps} distance are treated as dense-region points and represent normal operating conditions of the bridge. The samples that lie close to these dense regions but do not independently satisfy the density requirement are considered boundary points. All remaining samples that do not belong to any dense region are treated as outliers or anomalies.

A simplified illustration of the DBSCAN method~\cite{jaiswal2025leveraging} is presented in Fig.~\ref{dbscan_design}. As shown, six distinct groups are formed based on the clustering process, where the neighbourhood radius \textit{eps} defines the cluster boundaries. A group is identified as a valid cluster only if it contains at least \textit{minpts} data points. For instance, when \textit{minpts} is set to 3, Groups 1, 2, 4, and 6 satisfy the clustering criteria and are classified as clusters, while Groups 3 and 5 are treated as outliers. In contrast, increasing \textit{minpts} to 5 results in Groups 3, 5, and 6 being labelled as outliers.

\section{Proposed Anomaly Detection Model}
\label{prop_model}
\begin{figure} [t!]
  \centering
  \includegraphics[width=\columnwidth, height=5.25cm, keepaspectratio]{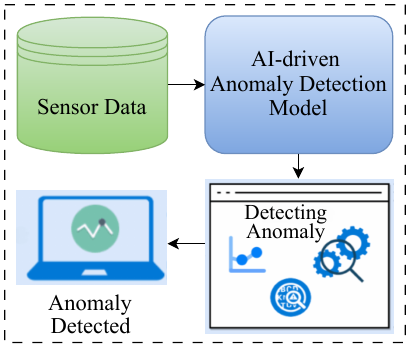}
  \caption{The proposed anomaly detection model.}
  \label{prop_ai_model} \vspace{-3mm}
\end{figure}

The primary goal of the proposed anomaly detection model is to accurately identify abnormal events for effective bridge monitoring. To achieve this goal, the proposed system architecture integrates multiple cooperating components, as illustrated in Fig.~\ref{prop_ai_model}. The architecture consists of three main stages. First, sensor data collected from the iBridge sensor device (see Section~\ref{data_coll}) are preprocessed and prepared as input to the anomaly detection model. Second, the AI-based anomaly detection model is trained using the processed data, during which it learns salient feature patterns. Finally, the trained model detects anomalous events and outputs anomaly predictions, which can be used to identify potential failures or accident-related incidents in the bridge monitoring system.

\section{Experimental Dataset}
\label{dset}
This section describes the sensor device used for data acquisition and the real-time bridge monitoring dataset.

\subsection{Sensor Data Acquisition  Device: iBridge}
\label{data_coll}
\begin{figure} [t!]
  \centering
  \includegraphics[width=\columnwidth, height=5.0cm, keepaspectratio]{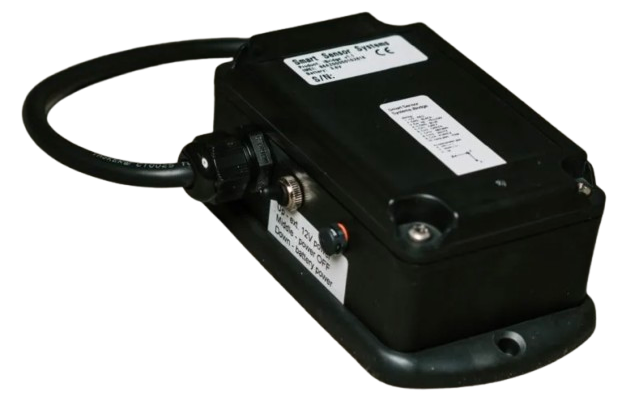}
  \caption{A simple illustration of the iBridge sensor device.}
  \label{ibridge_device} 
\end{figure}

The bridge monitoring data are collected using a sensor-based device, iBridge, as illustrated in Fig.~\ref{ibridge_device}. The iBridge~\cite{ibridge_sss} is a compact and smart device equipped with a multifunctional sensor-based system that transmits sensor data to a cloud platform via 4G communication. It supports battery operation and can be interfaced with a wide range of sensors and transmitters. The iBridge is specifically designed for use with strain gauges and can measure various structural responses, such as acceleration, tensile and compressive stresses, cracking, displacement of mass and collapse, strain on the metal beams, weight, pressure, vibration, angular variation, torque, temperature, liquid level, and moisture.

\begin{figure} [t!]
  \centering
  \includegraphics[width=0.98\columnwidth, height=6.0cm,keepaspectratio]{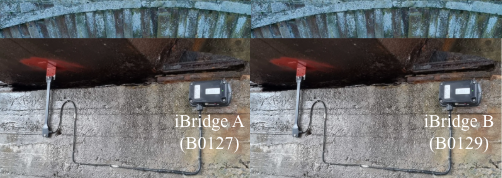}
  \caption{Illustration of installed iBridge devices on the bridge.}
  \label{ibridge_install} \vspace{-4mm}
\end{figure}

The iBridge device incorporates an integrated accelerometer and temperature sensor, functioning as a multifunctional data logger. It is easy to install on various materials and surfaces, can be relocated and reused, and supports adaptable battery configurations based on deployment needs. The direct sensor connectivity enables cable-free installation and no power connection. The device is connected to an intelligent cloud-based platform for configuration and data storage, providing continuous 24/7 access to monitoring data or seamless integration with the enterprise resource planning (ERP) system.

The iBridge device is suitable for a wide range of monitoring applications. For example, in the building and construction sector, it can be used to monitor cracks, settlement-related damages, vibrations, and moisture. In transportation infrastructure, including roads, ports, bridges, and railways, it can enable the assessment of changes in load-bearing capacity, overload conditions, structural forces, and unforeseen incidents. 

\subsection{Bridge Monitoring Dataset}
\label{bridge_data}
The bridge monitoring data in this study are collected from a bridge located in Inland, Norway. Two iBridge sensor devices (iBridge~A and iBridge~B) are installed beneath the bridge, approximately 70 meters apart, as shown in Fig.~\ref{ibridge_install}. For accident detection, real-time sensor data from both devices is considered over eleven days, from August 15 to August 25, 2025. The sensor data includes measurements of acceleration (in mG) in X, Y and Z directions, respectively, and strain on the metal beams (in mV). It contains a total of 3,775,112 samples. These measurements consist of numerical values. The data is sampled at a frequency of 5 Hz, corresponding to one measurement every 0.2 seconds.

\vspace{-2mm}
\subsection{Data Preprocessing}
\label{data_prepro}
The collected raw sensor data contain inconsistencies and missing values, which can adversely affect accuracy and model performance. Therefore, the dataset is preprocessed to enhance data quality, reliability, and suitability for ML models analysis. The preprocessing stage involves data cleaning and handling missing entries. As the number of missing values is limited, these samples are discarded from the dataset. Alternatively, transfer learning~\cite{jaiswal2023location} techniques can be employed to address missing data or data scarcity challenges. The processed dataset is then used to train ML models for anomaly detection. 

\begin{figure} [h!]
    \centering
    \includegraphics[width=\columnwidth, height=5.25cm,keepaspectratio]{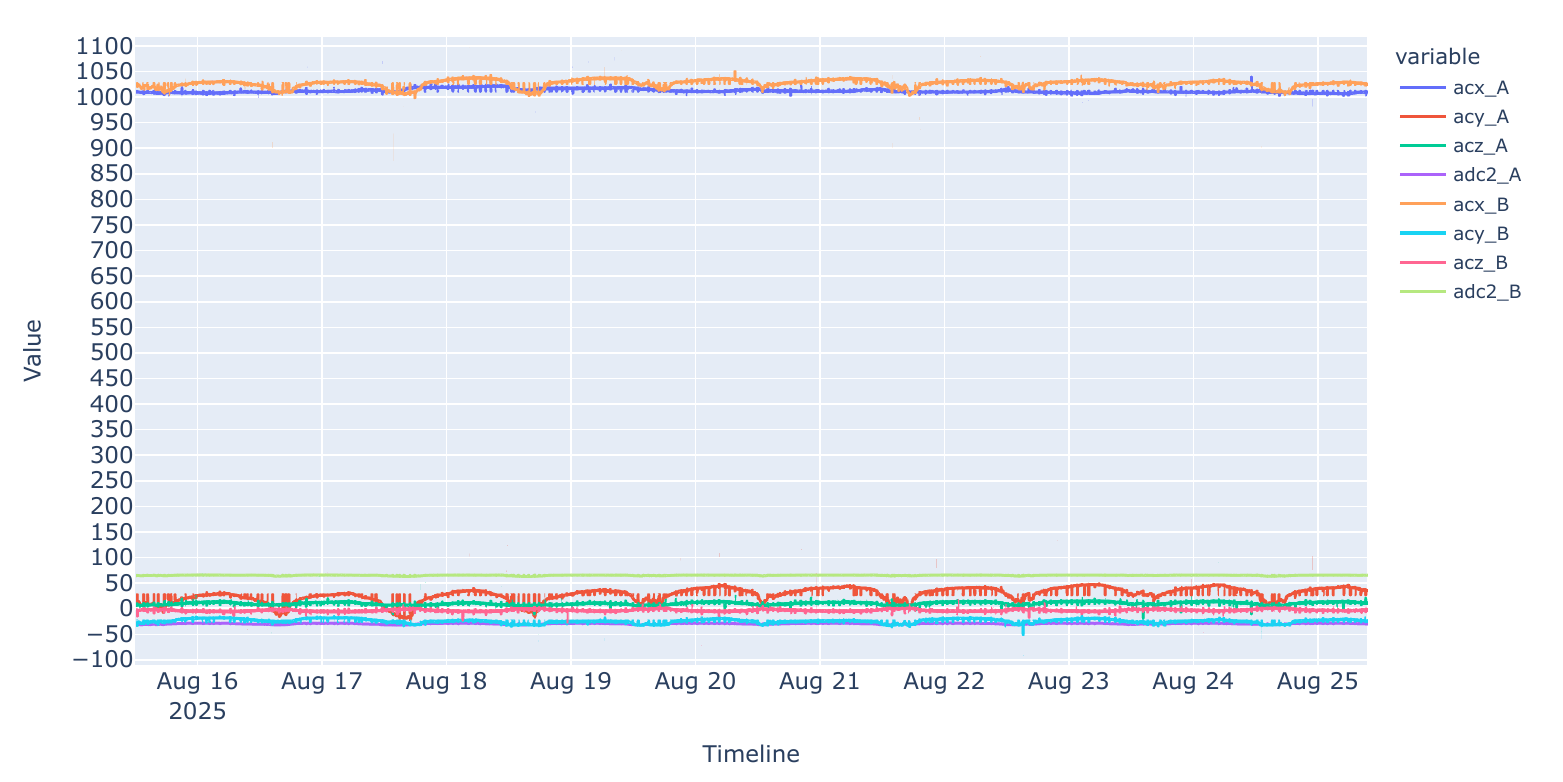}
    \caption{All features in the bridge monitoring dataset.}
    \label{final_ibridge_data} \vspace{-3mm}
\end{figure}

Next, we denote the notations for the features corresponding to iBridge A and iBridge B. For iBridge A, the accelerations in X, Y, Z directions, and strain
on the metal beams are denoted as $\text{acx\_A}$, $\text{acy\_A}$, $\text{acz\_A}$, and $\text{adc2\_A}$, respectively. Similarly, for iBridge B, the corresponding features are denoted as $\text{acx\_B}$, $\text{acy\_B}$, $\text{acz\_B}$, and $\text{adc2\_B}$, respectively. Since five samples are recorded per second, the data are resampled at a one-minute interval to enhance data pattern clarity, highlight anomaly peaks, improve temporal uniformity, and stabilize model learning, thereby enabling more reliable ML models. For visualization, the final resulting dataset is shown in Fig.~\ref{final_ibridge_data}. 

\begin{figure} [h!]
    \centering
    \includegraphics[width=\columnwidth, height=5.25cm,keepaspectratio]{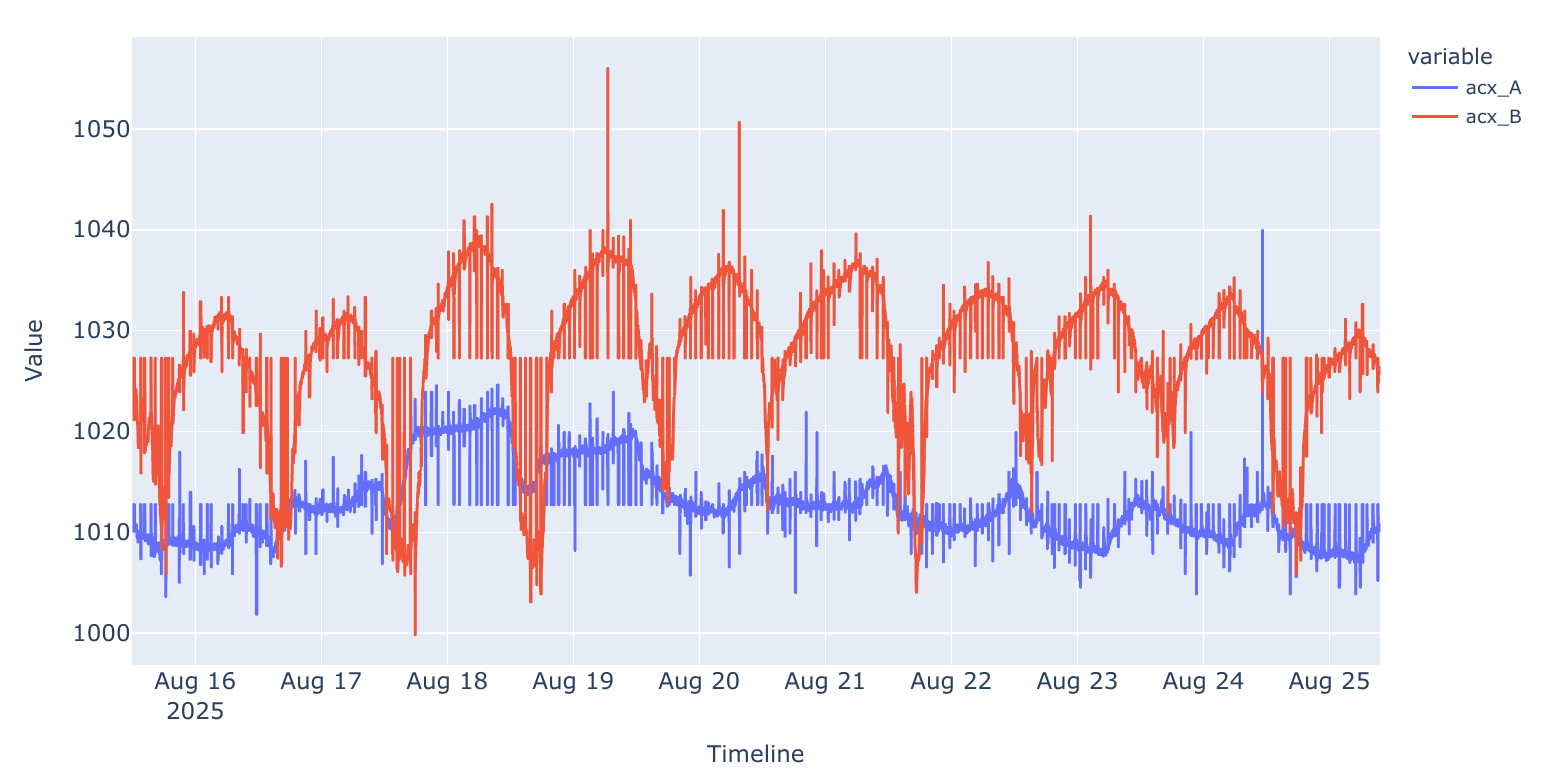}
    \caption{$\text{acx\_A}$ and $\text{acx\_B}$ of the bridge monitoring dataset.}
    \label{ibridge_data_acx_acy} \vspace{-3mm}
\end{figure}

\begin{figure} [t!]
    \centering
    \includegraphics[width=\columnwidth, height=5.5cm,keepaspectratio]{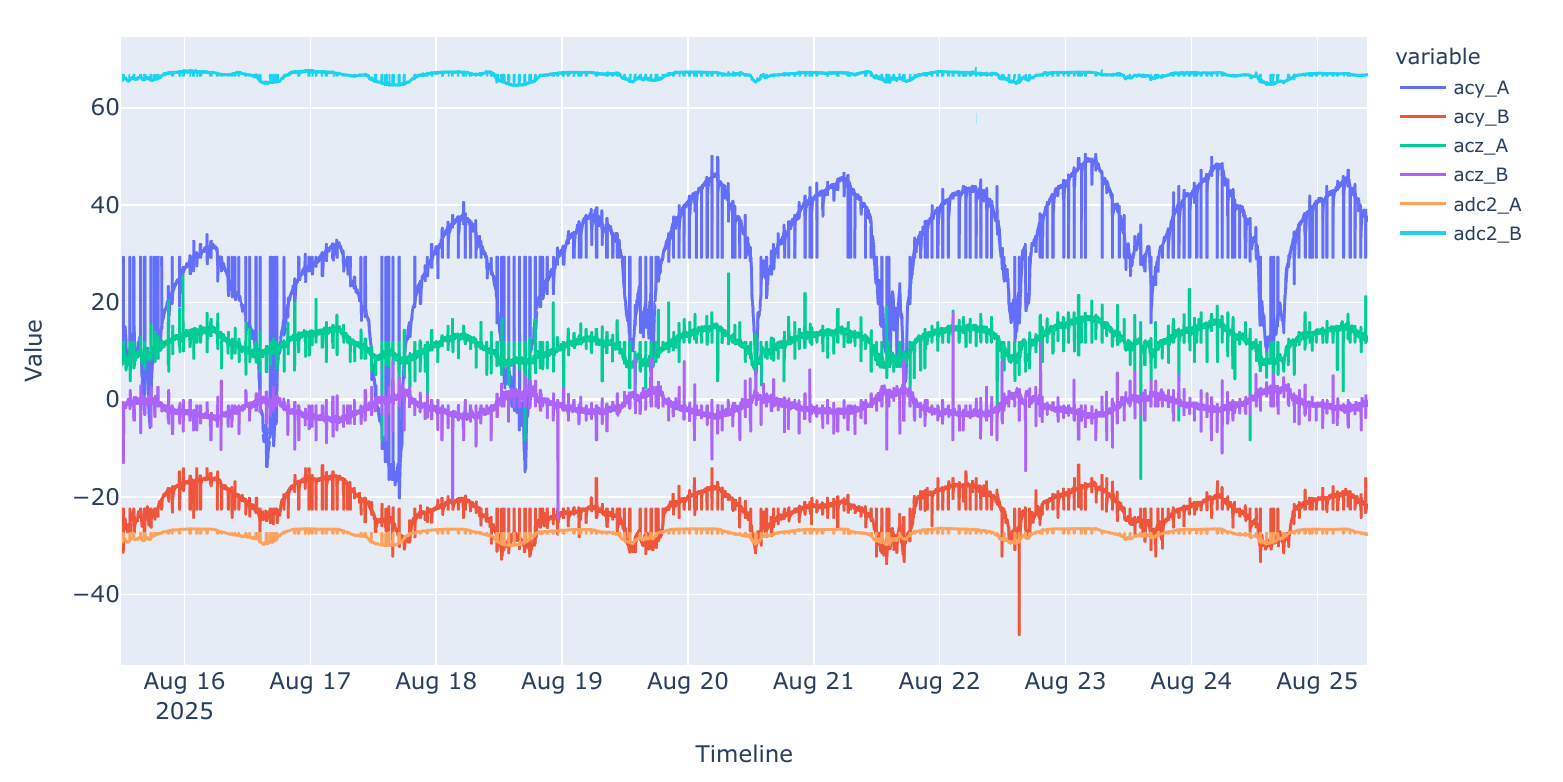}
    \caption{All features except $\text{acx\_A}$ and $\text{acx\_B}$ of the dataset.}
    \label{ibridge_data_except_acx_acy} 
\end{figure}

Since the data ranges of $\text{acx\_A}$ and $\text{acx\_B}$ are significantly larger than those of the other features, they are illustrated separately in Fig.~\ref{ibridge_data_acx_acy}. The remaining features, excluding $\text{acx\_A}$ and $\text{acx\_B}$, are shown in Fig.~\ref{ibridge_data_except_acx_acy}. 
\vspace{-2mm}

\section{Results and Discussions} 
\label{res_dis}
This section presents the experimental environment and a comprehensive analysis of the results.

\subsection{Experimental Environment} 
\label{sys_set}
All algorithms are implemented in Python~3.13.6 using the Keras framework built on TensorFlow~2.2.0 and executed on a MacBook with an Apple M4 chip and 16~GB of RAM.

\subsection{Ground Truth}  
\label{sub_anal} 
\begin{figure} [t!]
    \centering
    \includegraphics[width=\columnwidth, height=5.5cm,keepaspectratio]{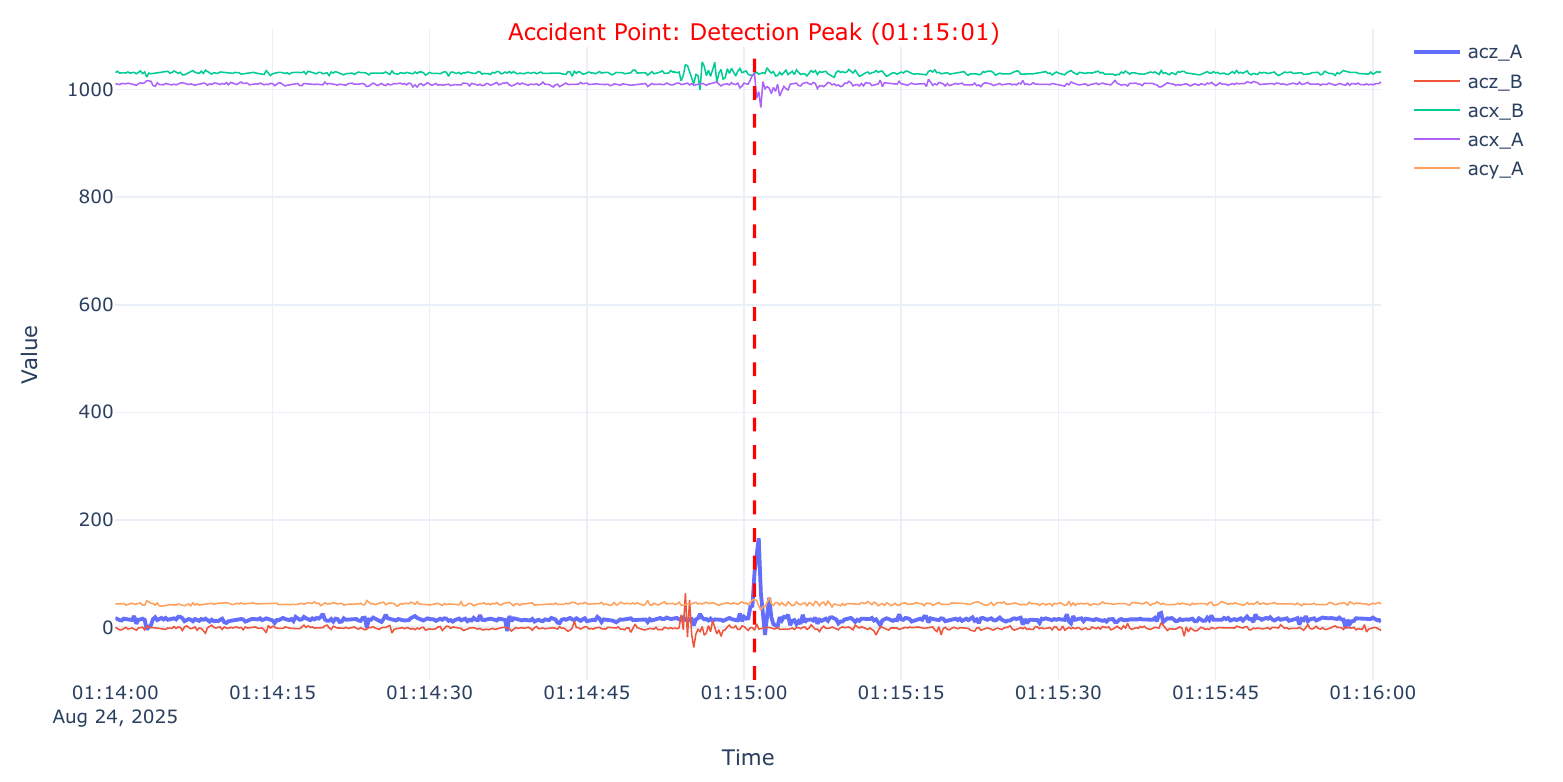}
    \caption{Actual (ground-truth) accident timestamp.}
    \label{act_acc_pnt} \vspace{-3mm}
\end{figure}

The exact timing of the accident is subjectively determined through a detailed manual analysis of the sensor data, which serves as the ground truth in this study, as shown in Fig.~\ref{act_acc_pnt}. The accident timing is identified as August~24, 2025, at 01:15~AM. Although this manual inspection enables accurate identification of the accident event, it requires expert knowledge and substantial human effort, making it impractical for continuous and real-time bridge monitoring. To address this limitation, we subsequently present ML-based anomaly detection models designed to automatically identify such events in a smart bridge monitoring system.

\subsection{Proposed Model Analysis} 
\label{ana_result} 
The bridge monitoring dataset described in Section~\ref{bridge_data} is employed to train three different ML models (see Section~\ref{ml_bgd}) for detecting the precise timing of the accident as an anomaly in the smart bridge monitoring system. The dataset is first scaled to a common range to ensure fair feature contribution and reduce false alarms. A grid search~\cite{elgeldawi2021hyperparameter} is then employed to obtain the optimal parameters for each model, which are presented in Table~\ref{para_used}.

\begin{table} [t!]
\caption{Parameters used in each ML model.}
\label{para_used} 
\centering
{
\setlength\tabcolsep{2.0pt}
\begin{tabular}{|c| c|}
\hline
Models & Parameters \\
\hline
Isolation Forest & $\text{n\_estimators}$\tablefootnote{In Isolation Forest, $\text{n\_estimators}$ is the number of isolation trees, contamination is the expected anomaly proportion and $\text{n\_jobs}$ is the available CPU cores (\text{n\_jobs}=-1= all available).}=1000, contamination=0.0001, n\_jobs=-1 \\
\hline
Autoencoder & input layer dimension=8, \\ 
& encoder input dimension=4, activation function=tanh, \\
& bottleneck layer dimension=1, activation function=linear, \\ 
& decoder input dimension=4, activation function=tanh, \\
& output layer dimension=8, activation function=linear, \\ & optimizer=adam, loss=mse, epochs=100, batch size=512 \\
\hline
DBSCAN & $\text{eps}$\tablefootnote{In DBSCAN, $\text{eps}$ is the maximum distance between two neighbour samples and $\text{min\_samples}$ is the minimum number of samples in a neighbourhood to form a dense region.}=0.8, $\text{min\_samples}$=3, $\text{n\_neighbors}$=2 \\
\hline
\end{tabular} 
}
\end{table}

\subsubsection{Isolation Forest Analysis} 
\label{decision_if} 
\begin{figure} [t!]
    \centering
    \includegraphics[width=\columnwidth, height=5.5cm,keepaspectratio]{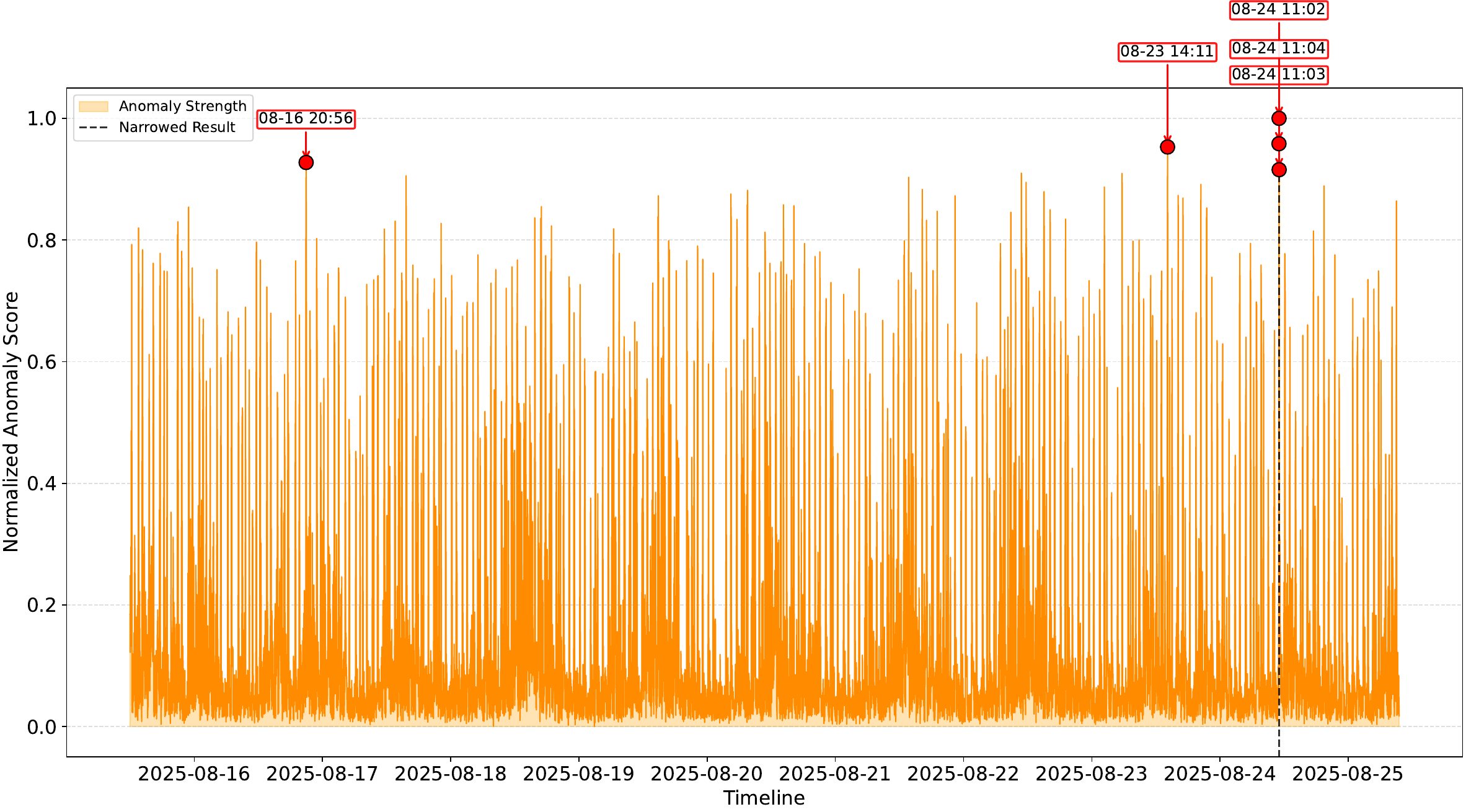}
    \caption{Detected anomalies using the Isolation Forest model.}
    \label{plot_if} \vspace{-3mm}
\end{figure}

The Isolation Forest model is trained with the parameters presented in Table~\ref{para_used}. A ranking-based anomaly selection strategy is employed to identify the most anomalous samples. The samples are ranked according to their normalized anomaly scores, where lower values indicate stronger anomalies and higher values correspond to more normal behaviour. This approach focuses on detecting the most extreme deviations. Based on this strategy, Fig.~\ref{plot_if} illustrates the detected anomalies. It can be seen that although a small number of prominent anomaly peaks (five in total) are observed, none coincide with the accident timestamp (August~24, 01:15~AM). This indicates that the Isolation Forest model cannot accurately determine the exact timing of the accident.

\subsubsection{Autoencoder Analysis} 
\label{decision_ae} 
\begin{figure} [t!]
    \centering
    \includegraphics[width=\columnwidth, height=5.0cm,keepaspectratio]{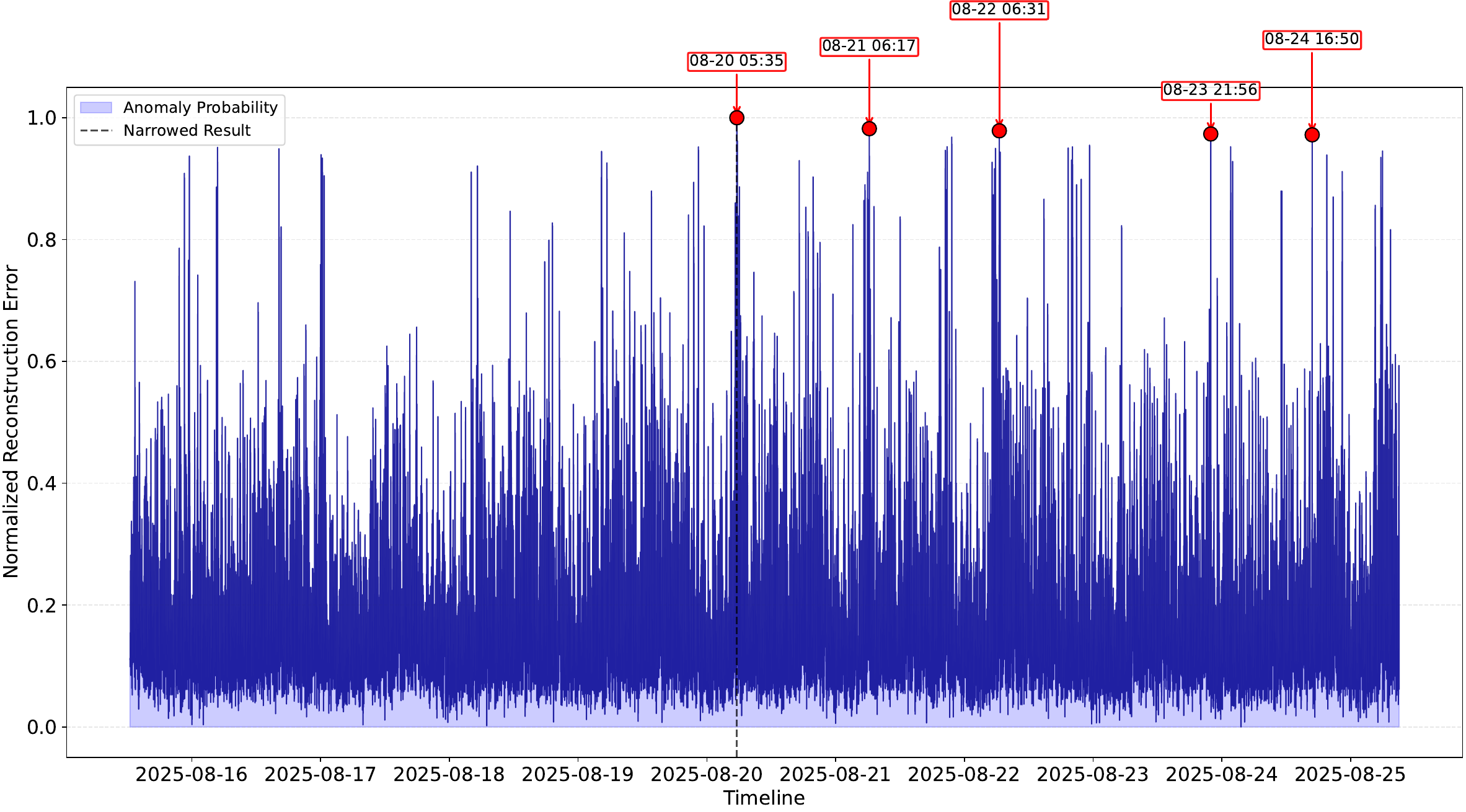}
    \caption{Detected anomalies using the autoencoder model.}
    \label{plot_ae} \vspace{-3mm}
\end{figure}

The autoencoder model is trained using the parameters presented in Table~\ref{para_used} and learns to reconstruct input samples based on the underlying data patterns. The reconstruction error (see Section~\ref{ae_algo}) is computed for the samples and used as an anomaly score to identify abnormal behaviour. The samples are subsequently ranked according to their normalized reconstruction error, where smaller values indicate accurate reconstruction and normal behaviour, while larger values correspond to poor reconstruction and anomalous behaviour. Based on this methodology, Fig.~\ref{plot_ae} presents the detected anomalies. It can be seen that although a limited number of anomaly peaks (five in total) are observed, none coincide with the accident timestamp (August~24, 01:15~AM). This suggests that the autoencoder model cannot accurately determine the exact timing of the accident.

\subsubsection{DBSCAN Analysis} 
\label{decision_dbscan} 
\begin{figure} [t!]
    \centering
    \includegraphics[width=\columnwidth, height=5.0cm,keepaspectratio]{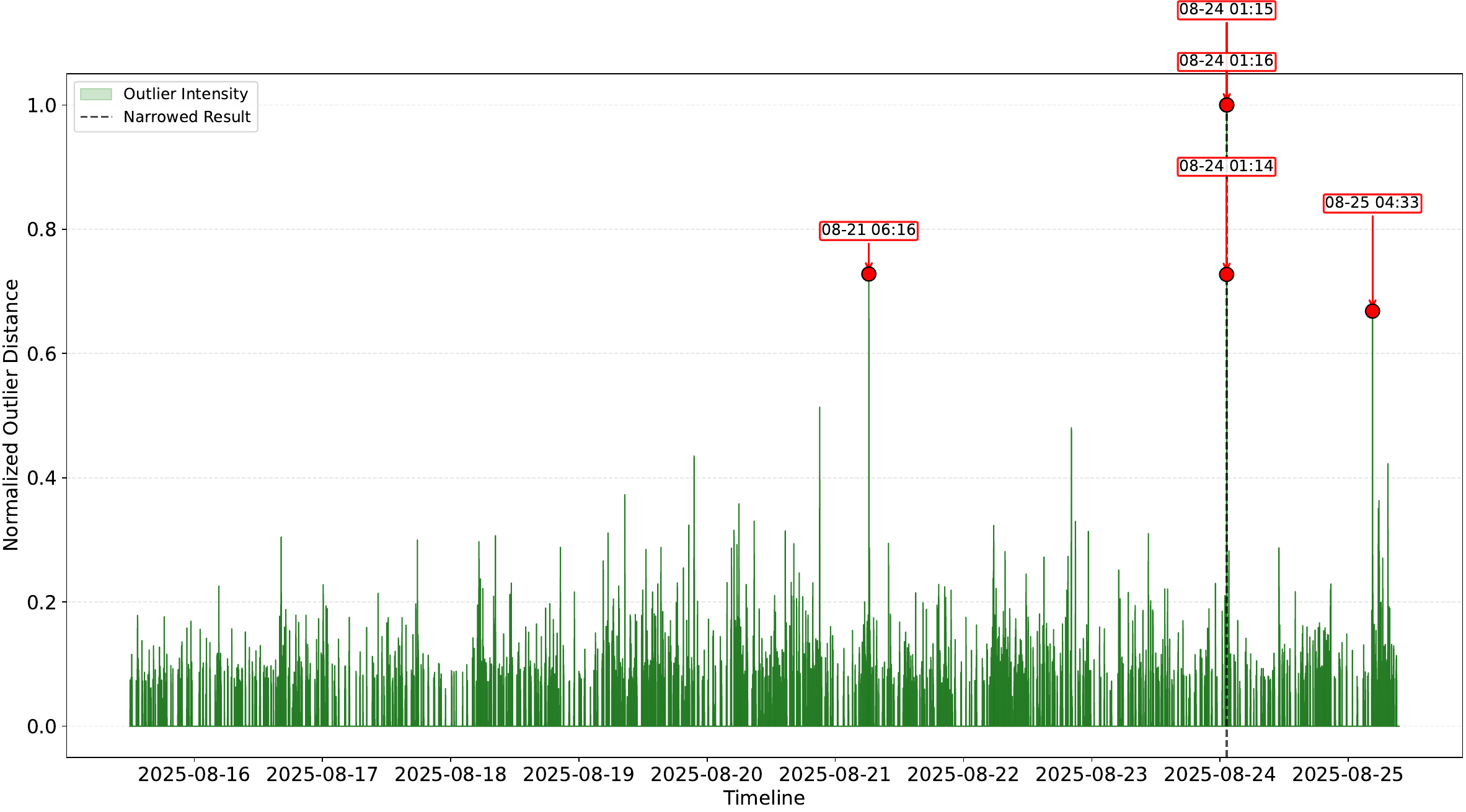}
    \caption{Detected anomalies using the DBSCAN model.}
    \label{plot_dbscan} \vspace{-3mm}
\end{figure}

The DBSCAN model is trained using the parameters presented in Table~\ref{para_used}. To identify anomalous samples, a nearest-neighbour distance-based anomaly selection strategy is adopted (see Section~\ref{dbscan_algo}). The samples are classified as normal or anomalous based on their normalized outlier distances (anomaly scores), where larger distances indicate stronger anomalies and smaller distances correspond to weaker anomalies. Using this approach, Fig.~\ref{plot_dbscan} shows the detected anomalies. Notably, a pronounced anomaly peak is observed at August~24, 01:15~AM, coinciding with the actual accident timestamp. This indicates that the DBSCAN model performs better than the other models and is capable of accurately detecting the timing of the accident, making it suitable for anomaly detection in the smart bridge monitoring system. 

\section{Conclusions and Future Work} 
\label{cons}
This paper investigates the anomaly detection problem in bridge monitoring systems using real-world sensor data collected by the iBridge sensor devices. The dataset is first analyzed and preprocessed to ensure data quality and reliability. Motivated by recent advances in artificial intelligence, we propose a simple machine learning model that learns discriminative features directly from sensor data and employs a decision module to detect anomalies, with a particular focus on identifying a bridge accident in Norway. Experimental results show that the proposed approach based on DBSCAN accurately detects anomalous events (bridge accident) and outperforms other evaluated ML models. Our proposed model has the potential to significantly enhance anomaly detection capabilities in bridge monitoring systems, thereby reducing the risk of catastrophic failures or accidents and improving public safety. Future work will explore the integration of more complex feature representations and the development of an early warning system for proactive smart bridge monitoring.

\section*{Acknowledgement}
This work was supported by Smart Sensor Systems AS through data provision and conference dissemination support.

\bibliographystyle{IEEEtran}
\bibliography{references}

\end{document}